\documentclass[runningheads,a4paper]{llncs}

\usepackage{amssymb}
\usepackage{graphicx}

\usepackage{url}
\newcommand{\keywords}[1]{\par\addvspace\baselineskip
\noindent\keywordname\enspace\ignorespaces#1}

\usepackage[colorlinks=true]{hyperref}
\usepackage{amsmath}
\usepackage{color}
\usepackage{subcaption}
\usepackage{enumitem}
\setlist{nolistsep}

\newcommand{\diag}{\operatorname{diag}}

\newcommand{\x}{\mathbf{x}}
\newcommand{\z}{\mathbf{z}}

\newcommand{\w}{\mathbf{w}}

\newcommand{\bvec}{\mathbf{b}}

\newcommand{\what}{\mathbf{\hat{w}}}
\newcommand{\xhat}{\mathbf{\hat{x}}}

\newcommand{\SMat}{\mathbf{S}}

\newcommand{\h}{\mathbf{h}}
\newcommand{\hhat}{\mathbf{\hat{h}}}
\newcommand{\zhat}{\mathbf{\hat{z}}}
\newcommand{\SHat}{\mathbf{\hat{S}}}
\newcommand{\g}{\mathbf{g}}

\newcommand{\f}{\mathbf{f}}

\newcommand{\M}{\mathbf{M}}

\newcommand{\bigo}{\mathcal{O}}

\newcommand{\xbar}{\mathbf{\bar{x}}}

\begin{document}
\mainmatter

\title{Learning Detectors Quickly Using \\ Structured Covariance Matrices}
\titlerunning{Learning Detectors Quickly Using Structured Covariance Matrices}

\author{Jack Valmadre$^{\ast \dagger}$\and Sridha Sridharan$^{\ast}$ \and Simon Lucey$^{\dagger}$}
\authorrunning{Learning Detectors Quickly Using Structured Covariance Matrices}

\institute{
  $^{\ast}$Queensland University of Technology, Brisbane, Australia \\
  $^{\dagger}$Commonwealth Scientific and Industrial Research Organisation, Australia \\
  \texttt{\{j.valmadre, s.sridharan\}@qut.edu.au, simon.lucey@csiro.au}
}

\maketitle

\begin{abstract}
Computer vision is increasingly becoming interested in the rapid estimation of object detectors.
Canonical hard negative mining strategies are slow as they require multiple passes of the large negative training set.
Recent work has demonstrated that if the distribution of negative examples is assumed to be stationary, then Linear Discriminant Analysis (LDA) can learn comparable detectors without ever revisiting the negative set.
Even with this insight, however, the time to learn a single object detector can still be on the order of tens of seconds on a modern desktop computer.
This paper proposes to leverage the resulting structured covariance matrix to obtain detectors with identical performance in orders of magnitude less time and memory.
We elucidate an important connection to the correlation filter literature, demonstrating that these can also be trained without ever revisiting the negative set.
\end{abstract}

\section{Introduction}

An issue of increasing importance in vision is the \emph{rapid} estimation of object detectors.
Historically, the central emphasis for learning a detector was the evaluation time, not the time it took to learn the detector.
For example, when learning a pedestrian detector it makes little practical difference if the detector takes a second, an hour, or even a day to estimate since the estimation task only needs to be accomplished once.
As long as detection performance is high and the evaluation efficient, training time has been considered of minimal consequence.

Computer vision is now, however, moving into tasks where the rapid estimation of well performing object detectors is becoming critical.
This is due not only to the arrival of datasets with tens or hundreds of thousands of classes (e.g.\ \cite{Deng2009,Dean2013}), but also to the use of the linear classifier as an elementary component in more complex systems.
The Exemplar SVM paradigm~\cite{Malisiewicz2011} involves training one linear detector per example to obtain more informative and expressive models.
Linear detectors are also used within algorithms to discover structure in weakly supervised or unsupervised datasets, such as human detection and pose estimation using poselets~\cite{Bourdev2009} and mid-level discriminative patch discovery~\cite{Singh2012}.
The computation of these tasks is controlled by the time it takes to generate a new detector.
Other scenarios which can clearly benefit from fast, lightweight algorithms include online learning, such as the Predator tracker~\cite{Kalal2009}, and mobile applications.

In this paper we discuss the limitations of current strategies to employ a large negative training set when learning object detectors.
Further, we present a generalised approach which enforces either Toeplitz or circulant structure in the second order statistics, and examine fast algorithms for both cases.

\subsection{Hard negative mining}

One of the fundamental questions when using machine learning to train a classifier for object detection is: how to treat the enormous negative set?
Any image which does not contain the object can contribute all of its sub-windows as valid negatives, quickly generating a myriad of examples.
Linear Support Vector Machines (SVMs) have been particularly useful in this regard, as they seek solutions that are sparse with respect to the training set (i.e.\ support vectors).
If these support vectors were known a priori, then only this smaller subset of the training set is required to find the optimal solution.
Finding this set is, however, no easier than solving the original problem.

Hard Negative Mining (HNM) has become extremely popular in vision literature over the last two decades~\cite{Dalal2005,Felzenszwalb2010Object}.
It takes advantage of the heuristic that the ``hardest'' examples should be support vectors.
Instead of training an SVM across the entire set, one instead maintains a subset of these difficult examples.
A random subset is used to learn the initial detector.
Each round uses the previous detector to exhaustively search the training set, incorporating the strongest false positives into the current active set for the next detector.
Besides its sparse treatment of the examples, the SVM is an attractive choice because it minimises the empirical classification loss, maximises the margin of the decision plane, is robust to outliers and corresponds to a convex minimisation problem.

\subsection{Linear Discriminant Analysis}

While HNM has proved to be effective, it is intrinsically unsuited to rapid computation of object detectors using large negative datasets.
For each new object class we want to learn, the negative set must be re-trawled, possibly multiple times, requiring access to the explicit examples in the full negative set.

In contrast, for Linear Discriminant Analysis (LDA), it is sufficient to summarise the negative set into its covariance and mean.
The parameters~$\w$ of the decision hyperplane~$\w^{T} \x = c$ are learnt by solving the system of equations
\begin{equation}
\SMat \w = \bvec \label{eq:lda}
\end{equation}
where $\SMat$ is the shared covariance of both classes and $\bvec = \xbar_{\text{pos}} - \xbar_{\text{neg}}$ is the difference between class means.
Computing the covariance matrix of the set of all sub-windows in many images is computationally infeasible if we naively treat it as a general set of vectors.
However, Hariharan et al.~\cite{Hariharan2012} recently highlighted the fact that this set follows a stationary distribution.
This implies that the covariance of two pixels is defined entirely by their relative displacement.
For time-series of length~$n$, this would manifest in an $n \times n$ symmetric Toeplitz covariance matrix with elements
\begin{equation}
S_{ij} = g[|j-i|] \label{eq:symm-toep}
\end{equation}
specified by a vector~$\g$ with only $n$ elements (see Figure~\ref{fig:unique}).
We will introduce a generalisation of the symmetric Toeplitz matrix for multi-channel images in the following section.
This succinct parameterisation
i) greatly reduces the time and memory required to gather and store covariance matrices, and
ii) is agnostic to the size of the signal window, enabling the construction of covariance matrices for different sizes after the fact.

Hariharan et al.~\cite{Hariharan2012} exploited this Toeplitz structure to efficiently obtain the second-order statistics.
To learn a detector, however, they still performed the onerous task of forming and factorising the full matrix.
This paper explores the utilisation of structured covariance matrices in this final stage to \emph{rapidly learn} detectors.
We entertain circulant as well as Toeplitz structure, illuminating an important connection to correlation filters~\cite{Mahalanobis1994}.

\section{Learning with Toeplitz structure}

\begin{figure}[t]
\centering
\includegraphics[width=120mm]{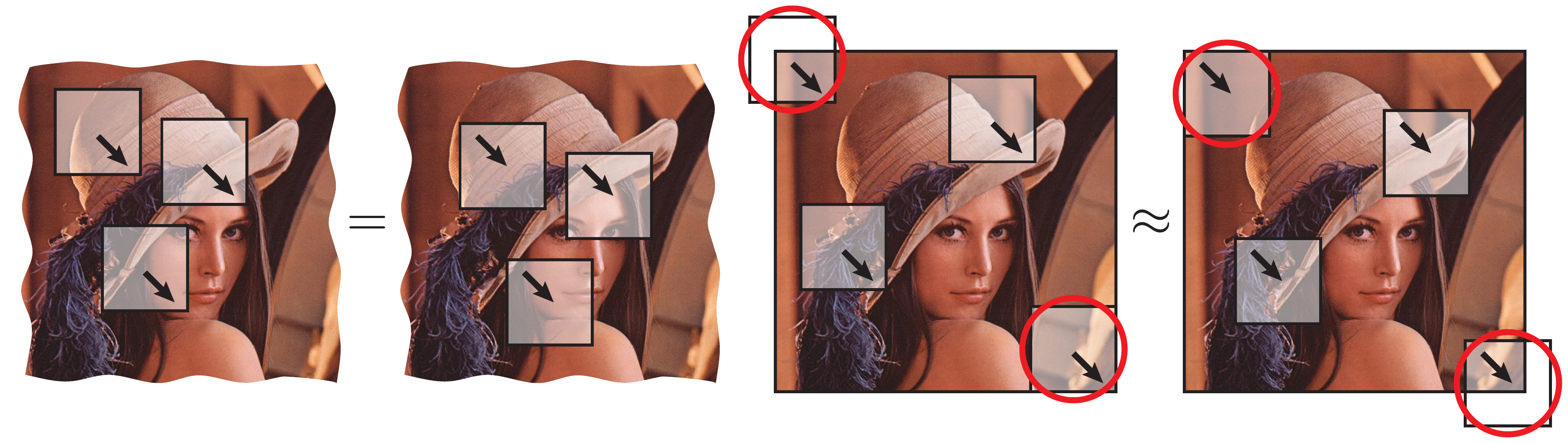}
\caption{
The covariance of all translated windows in an \emph{infinite} image (left) has exact Toeplitz structure since all pairs of pixels with a fixed relative displacement are observed in every position within the window.
However, the presence of boundaries in a finite image (right) perturbs the covariance matrix slightly from Toeplitz, as the location of the pixel pair within the window affects which of the border pixels it can observe (indicated by red circles).
Note that we only consider windows which lie wholly inside the image, without padding or extension of any kind.
Hariharan et al.~\cite{Hariharan2012} explicitly enforce this structure in LDA to greatly reduce computation and storage requirements when estimating the statistics.
}
\label{fig:inf-image}
\end{figure}

In fact, the covariance matrix of all translated windows in an image is only \emph{exactly} Toeplitz if the image has no boundaries.
This includes periodic images and images with infinite extent (refer to Figure~\ref{fig:inf-image}).
For such images, a translation of the underlying image does not affect the set of all sub-windows, and therefore the pair-wise statistics of two pixels is invariant to their absolute position.
However, the presence of finite image boundaries causes the matrix to diverge slightly from Toeplitz.
Hariharan et al.~\cite{Hariharan2012} explicitly enforce this structure on the covariance matrix.

In this section we introduce the concept of a ``block two-level Toeplitz'' matrix for multi-channel 2D signals (i.e.\ feature images), briefly review the formulation of LDA under stationarity and discuss methods for solving these systems.

\subsection{Block two-level Toeplitz matrices}

For time-series, the stationarity property implies the straightforward symmetric Toeplitz form in equation~\eqref{eq:symm-toep}.
However, we are primarily interested in applying this theory to feature images, which are multi-channel two-dimensional signals.
We therefore consider vectors~$\x$ which represent an $m \times n$ image patch with $k$ feature channels, whose elements are identified by $x_{p}[u, v]$ with $u \in \{0, \dots, m-1\}$, $v \in \{0, \dots, n-1\}$ and $p \in \{1, \dots, k\}$.
Note that we distinguish between ``channels,'' which do not have stationarity, and ``dimensions,'' which do.

\begin{figure}[t]
\centering
\begin{subfigure}[b]{0.5\textwidth}
  \centering
  \includegraphics[scale=0.35]{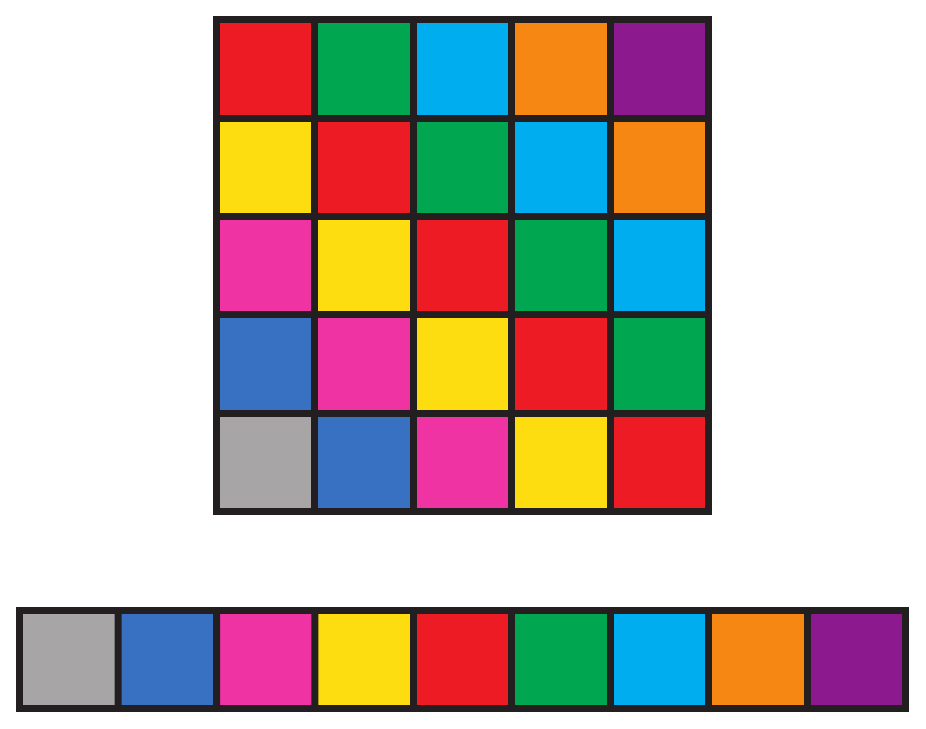}
  \hspace{1em}
  \includegraphics[scale=0.35]{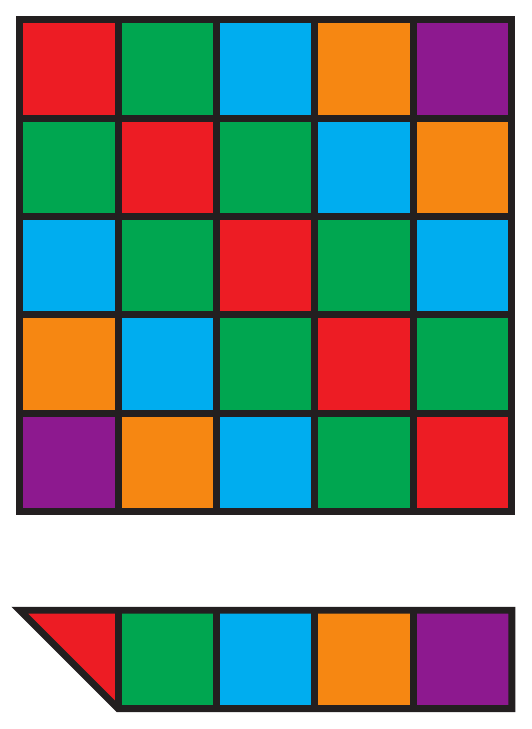}
  \caption{general/symmetric block Toeplitz}
\end{subfigure}
\hspace{1em}
\begin{subfigure}[b]{0.45\textwidth}
  \centering
  \includegraphics[scale=0.35]{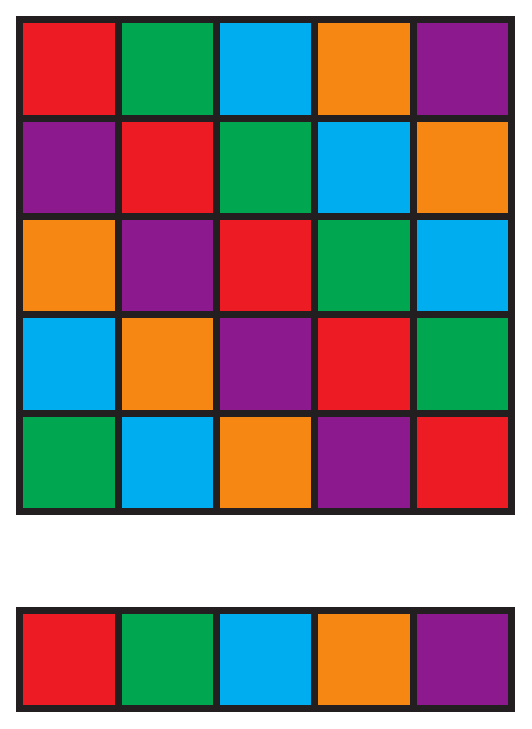}
  \hspace{1em}
  \includegraphics[scale=0.35]{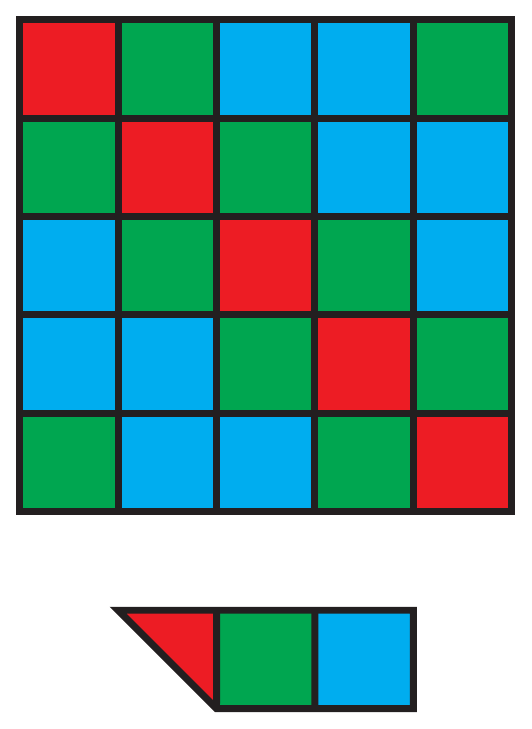}
  \caption{general/symmetric block circulant}
\end{subfigure}
\caption{
Block Toeplitz and circulant matrices shown beside their defining elements.
Circulant matrices are a subset of Toeplitz matrices whose elements re-appear on the left after disappearing on the right.
Any Toeplitz matrix can be embedded in a circulant matrix of roughly twice the size.
}
\label{fig:unique}
\end{figure}

Each element of the $mnk \times mnk$ matrix~$\SMat$ describes the covariance of channel~$p$ of some pixel~$(u, v)$ with channel~$q$ of some other $(i, j)$.
Enforcing two-dimensional stationarity constrains the elements of this matrix according to
\begin{equation}
S_{(u, v, p), (i, j, q)} = g_{pq}[i-u, j-v]
\end{equation}
where we use $(u, v, p)$ and $(i, j, q)$ to refer to indices into the vectorised multi-channel image whilst remaining agnostic to the specific order of vectorisation.
The array of relative displacement covariances~$\g$ is analogous to the first row of the symmetric Toeplitz matrix in equation~\eqref{eq:symm-toep} in that it captures the unique elements of the full matrix (see Figure~\ref{fig:unique}).
The symmetry of~$\SMat$ implies one mode of symmetric redundancy $g_{pq}[du, dv] = g_{qp}[-du, -dv]$ in the array.

\subsection{Direct methods for solving Toeplitz systems}

Hariharan et al.~\cite{Hariharan2012} assumed that the distribution of negative images sub-windows was stationary, restricting their covariance matrix to be block two-level Toeplitz.
Further, they assumed that the relatively few positive examples would contribute negligibly to the estimation of the shared covariance matrix~$\SMat$, and hence adopted the covariance matrix of the negative examples in equation~\eqref{eq:lda}.
Therefore this block two-level Toeplitz structure also exists in the system of equations which we now seek to solve, although this has not previously been taken advantage of.

It is well known that circulant matrices are diagonalised by the Fast Fourier Transform (FFT).
This enables both matrix-vector products and the solution of a linear system for any $n \times n$ circulant matrix to be computed in $\bigo(n \log n)$ time.
Multiplication of a vector by any $n \times n$ Toeplitz matrix can also thus be performed in~$\bigo(n\log{n})$ time by extending the original vector with zeros and multiplying it by the $(2n-1) \times (2n-1)$ circulant matrix which contains the Toeplitz matrix, taking only a subset of $n$ output elements.
Unfortunately, there is no analogous method to \emph{solve} a Toeplitz system.

There is, however, an extensive and varied body of literature surrounding this inverse problem, and we briefly review some key results here.
Recall that a general $n \times n$ system of equations can be factorised in $\bigo(n^{3})$ time with solutions then obtained in $\bigo(n^{2})$ time.
Levinson recursion~\cite{Levinson1947,Trench1964} allows Toeplitz systems to instead be factorised in $\bigo(n^{2})$ time, with the Gohberg-Semencul formula~\cite{Gohberg1972} enabling solutions to then be obtained in $\bigo(n\log{n})$ time.
This is entirely without inflicting the $\bigo(n^{2})$ memory requirement of instantiating the explicit matrix or its inverse.
There also exist ``superfast'' or ``asymptotic'' algorithms~\cite{Brent1980,Ammar1988} which solve a system in $\bigo(n \log^{2}{n})$ time without factorisation, although the hidden coefficients can be large.
Levinson recursion has been generalised to solve $nk \times nk$ block (one-level) Toeplitz systems, comprising an $n \times n$ Toeplitz structure of arbitrary $k \times k$ blocks, in an algorithm that takes $\bigo(n^{2} k^{3})$ time~\cite{Akaike1973}.

Unfortunately, in the extension to multi-level Toeplitz matrices, which are our primary interest in vision, the complexity of the specialised algorithms grows at the same rate as that of the general algorithms.
Wax and Kailath~\cite{Wax1983} compared two-level Toeplitz matrices to general block Toeplitz matrices and found merely that it's possible to improve the speed by a factor of two.
In fact, Yagle~\cite{Yagle2001} has argued that ``any fast algorithm for Toeplitz-block Toeplitz systems must be non-Levinson-like in nature.''
Some exceptions have been identified, such as where one of the levels is triangular~\cite{Turnes2012} or where the Toeplitz matrix contains only low frequencies~\cite{Yagle2001}.

\subsection{Iterative methods for solving Toeplitz systems}

Therefore we resort to iterative methods, coupled with the aforementioned fast routine to quickly compute matrix-vector products.
Critically, fast multiplication in the Fourier-domain \emph{does} extend to block multi-level Toeplitz matrices, easily confirmed in that $\z = \SMat \x$ implies
\begin{equation}
z_{p}[u, v]
\; = \; \sum_{i, j, q} S_{(u, v, p), (i, j, q)} \, x_{q}[i, j]
\; = \; \sum_{q} \sum_{i, j} g_{pq}[i-u, j-v] \, x_{q}[i, j] \enspace .
\label{eq:toep-mul}
\end{equation}
This amounts to a sum of two-dimensional cross-correlations, which can each be performed exactly and efficiently using the FFT with suitable zero-padding.
Using $d = mn$ to denote the number of pixels in the window, the above can be computed in $\bigo(k^{2}d\log{d})$ time.

In fact, a number of past works have proposed to solve plain (i.e.\ neither block nor multi-level) Toeplitz systems using the Conjugate Gradient (CG) method.
The convergence rate of CG depends on both the condition number of the matrix and how tightly clustered its eigenvalues are~\cite{Nocedal2006}.
The Preconditioned Conjugate Gradient (PCG) method instead solves the equivalent problem $\M \SMat \w = \M \bvec$, where the preconditioner $\M$ must be full rank and $\M \SMat$ has more desirable spectral properties than $\SMat$ alone.
Most works have centred around the choice of preconditioner, with Chan and Ng~\cite{RChan1996} in particular arguing that an effective preconditioner renders the number of iterations a small constant, yielding the solution to an $n \times n$ Toeplitz system in $\bigo(n\log{n})$ time.

The ideal choice is $\M = \SMat^{-1}$, however multiplying by this matrix corresponds to solving the original problem.
Circulant matrices make attractive preconditioners because they are easily inverted and are similar to (in fact a subset of) Toeplitz matrices.
Strang~\cite{Strang1986} originally proposed the inverse of the circulant matrix which is constructed by grafting the inner diagonals of a Toeplitz matrix into the outer opposite corners.
This was later shown to minimise the distance from the Toeplitz matrix under both the $L_{1}$ and $L_{\infty}$ operator norms, and guarantee superlinear convergence for a large class of problems~\cite{RChan1989Hermitian}.
Chan~\cite{TChan1988} instead proposed the nearest circulant matrix in the Frobenius sense and observed empirically that it was more effective at reducing the condition number and producing a clustered spectrum.

Circulant preconditioners are of particular interest for our problem, since it is known from the extension of correlation filters to multi-channel signals that block two-level circulant matrices are easily inverted.
This will be reviewed in Section~\ref{sec:circ}.
Two-level circulant preconditioners have previously been explored for block Toeplitz~\cite{TChan1994} and two-level Toeplitz systems~\cite{RChan1996}.
Serra Capizzano and Tyrtyshnikov~\cite{SerraCapizzano2000} presented the theoretical result that multi-level circulant preconditioners are not guaranteed superlinear convergence for multi-level Toeplitz matrices by the same mechanism, noting that fast convergence is still possible in practice.
Our final preconditioner is presented in Section~\ref{sec:circ} and an empirical comparison to other methods in Section~\ref{sec:results}.

An alternative strategy is to use the Alternating Direction Method of Multipliers to solve the padded circulant system with the additional linear constraint that the padding be zero as in~\cite{Bristow2013,Almeida2013}.
However, our initial experiments suggest this to converge slowly, perhaps because the ``filters'' in previous works have been compact, whereas ours have full support.

\subsection{Accumulating statistics in the Fourier domain}

While it is not the primary focus of this paper, we briefly mention that the FFT can be used to accelerate estimation of the stationary covariance matrix.
To estimate the covariance of pixels with relative displacement $(du, dv)$ from each image~$\f$, one computes for all $p, q \in \{1, \dots, k\}$ the normalised sum
\begin{equation}
g_{pq}[du, dv] = \left(\sum_{\f} \sum_{u, v} f_{p}[u, v] \, f_{q}[u+du, v+dv]\right) \Bigg/ \left(\sum_{\f} \sum_{u, v} 1\right)
\label{eq:rel-disp-covar}
\end{equation}
where all $u$ and $v$ are considered such that $(u, v)$ and $(u+du, v+dv)$ lie inside the image.
Note that each image~$\f$ may be of arbitrary size.
To construct a covariance matrix for patches of $m \times n$ pixels, the elements of~$g_{pq}[du, dv]$ must be known for $du \in \{-m+1, \dots, m-1\}$ and $dv \in \{-n+1, \dots, n-1\}$.
The covariance could be set to zero beyond a certain horizon, giving a band-Toeplitz system, although we do not consider it in this work.
It's not necessary to subtract the mean of each element during estimation, as this can be done later using
\begin{equation}
\frac{1}{N} \sum_{\x} (\x - \xbar) (\x - \xbar)^{T} \; = \; \frac{1}{N} \sum_{\x} \x \x^{T} - \xbar \xbar^{T} \enspace .
\end{equation}
The mean~$\xbar$ of the stationary distribution is uniformly constant per channel with elements $\bar{x}_{p}[u, v] = \mu_{p}$.

Consider computing this for a single large image of size $M \times N$ and let $D = M N$.
To naively gather the general (non-Toeplitz) covariance matrix from all windows in this image would require $\bigo(k^{2} d^{2} D)$ time.
Hariharan et al.~\cite{Hariharan2012} instead compute the relative displacement covariance via equation~\eqref{eq:rel-disp-covar} in $\bigo(k^{2} d D)$ time.
Recognising the numerator as cross-correlation (per channel pair), it is possible to obtain the covariance from a single image in $\bigo(k^{2} D\log{D})$ time.
Not only is this much faster, but it enables statistics to be gathered for templates as large as the image itself at no additional asymptotic cost.
This contribution runs orthogonal to the other contributions of the paper, which concentrate on learning a detector having already obtained these statistics.

\section{Learning with circulant structure}
\label{sec:circ}

Circulant matrices are a subset of Toeplitz matrices which further satisfy
\begin{equation}
S_{ij} = h[(j-i) \bmod n] \enspace ,
\end{equation}
with an $n \times n$ matrix entirely defined by the vector~$\h$ of length~$n$, which has only $\lceil n/2 \rceil$ unique elements if the matrix is symmetric (see Figure~\ref{fig:unique}).
Whereas the Toeplitz constraint followed intuitively from the stationarity property, the motivation to consider circulant matrices is purely computational.
All circulant matrices are diagonalised by the FFT, making it possible to solve circulant systems in $\bigo(n\log{n})$ time.
For comparison, recall that classical methods to solve Toeplitz systems take $\bigo(n^{2})$ and ``superfast'' methods $\bigo(n\log^{2}{n})$ time.
This disparity grows further still when considering multi-level structure.

In this section, we introduce the block two-level circulant matrix which arises in Multi-Channel Correlation Filters~\cite{Henriques2013} and then develop a method to derive such a matrix from the Toeplitz covariance matrix rather than by sampling numerous image patches.
We propose the use of the circulant system either directly or as a preconditioner for the Toeplitz problem.

\subsection{Block two-level circulant matrices}

Generalising from time-series to multi-channel images, the covariance matrix has block two-level circulant structure
\begin{equation}
S_{(u, v, p), (i, j, q)} = h_{pq}[(i-u) \bmod m, (j-v) \bmod n] \enspace .
\label{eq:circ}
\end{equation}
As detailed in Henriques et al.~\cite{Henriques2013}, this matrix is \emph{block}-diagonalised by taking the two-dimensional Fourier transform of each channel independently.
This is observed in that multiplication is now equivalent to \emph{periodic} cross-correlation, with $\z = \SMat \w$ implying
\begin{equation}
z_{p}[u, v] = \sum_{q} \sum_{i, j} h_{pq}[(i-u) \bmod m, (j-v) \bmod n] \, x_{q}[i, j] \enspace .
\end{equation}
Unlike equation~\eqref{eq:toep-mul}, the presence of the modulo operators permits us to express this directly in the Fourier domain, importantly without zero-padding
\begin{equation}
\zhat_{p} = \sum_{q} \diag\bigl(\hhat_{pq}\bigr)^{*} \, \xhat_{q} \enspace .
\end{equation}
Here we use $\x_{q}$ to represent each single-channel ``plane'' of the multi-channel signal, $\h_{pq}$ for the pair-wise channel planes which define the circulant matrix, $\xhat = \mathcal{F} \x$ as short-hand for the Fourier transform of~$\x$, and $\mathbf{A}^{\ast}$ to denote Hermitian transpose.
The above can be re-arranged into independent systems per pixel
\begin{equation}
\zhat[u, v] = \SHat_{uv} \, \what[u, v]
\end{equation}
where $\x[u, v]$ gives the sample at pixel $(u, v)$ as a $k$-dimensional vector, and each $k \times k$ block is defined
\begin{equation}
\SHat_{uv} = \bigg( \hat{h}_{pq}^{*}[u, v] \bigg)_{pq} \enspace .
\end{equation}

Let $d = mn$ denote the number of pixels.
It takes $\bigo(k^{2}d\log{d})$ time to compute the transforms of the channel-pair slices~$\h_{pq}$ and $\bigo(k^{3}d)$ time to factorise the individual blocks~$\SMat_{uv}$.
To then obtain a solution requires $\bigo(k^{2}d)$ time to solve the factorised systems and $\bigo(kd\log{d})$ time for transforms.
This is the computational device which enables the technique of Multi-Channel Correlation Filters (MCCFs)~\cite{Boddeti2013,Henriques2013,KianiGaloogahi2013}.

\subsection{Correlation filters}

\begin{figure}[t]
\centering
\includegraphics[height=35mm]{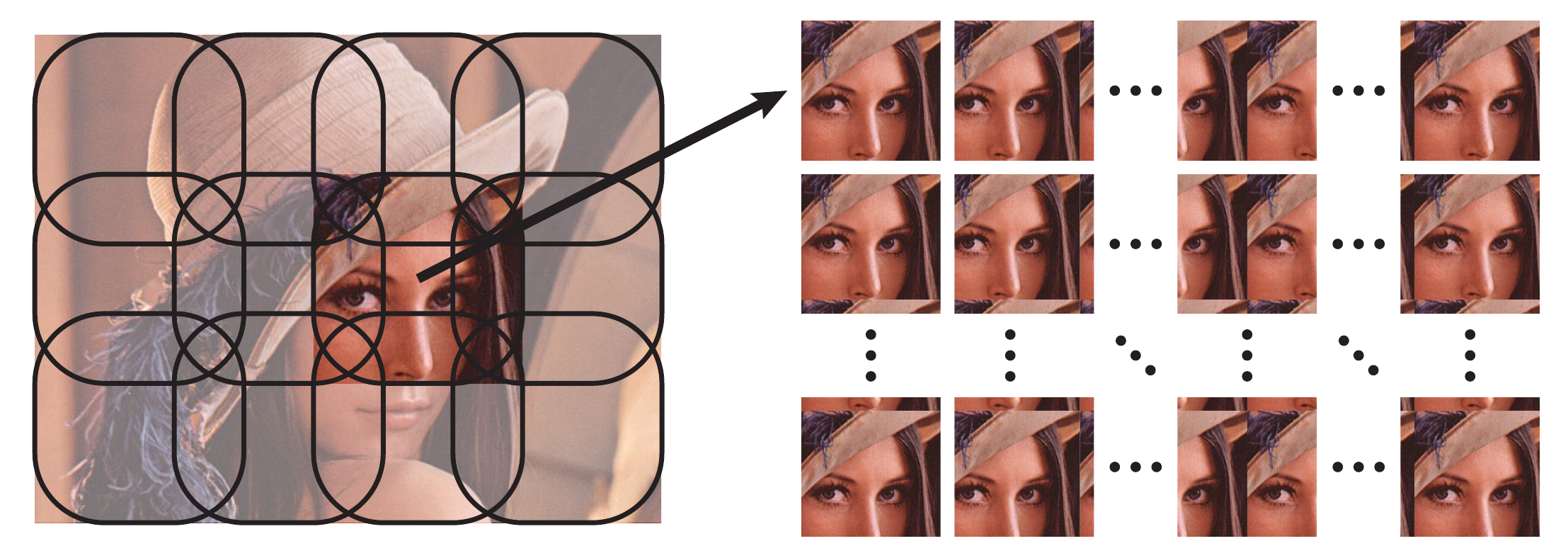}
\caption{
Using correlation filters, Henriques et al.\ approximate the set of all translated windows in an image with all circular shifts (right) of a coarsely-sampled set of windows which cover the image (left).
This results in a generalised circulant matrix which can be inverted in closed form.
Rounded rectangles illustrate overlap.
}
\label{fig:mccf-hnm}
\end{figure}

Circulant covariance matrices result from sets of signals which comprise all circular shifts of every member of a base set.
Correlation filters~\cite{Mahalanobis1994} use this to simultaneously introduce \emph{more} examples and make the problem \emph{easier} to solve, although the additional examples are not necessarily helpful as they in fact serve to constrain the problem.
For single-channel signals (typically greyscale images), the diagonalised matrix is obtained directly from the Fourier transform of the examples in the base set
\begin{equation}
\SHat = \sum_{\x} \diag(\xhat) \diag(\xhat)^{*} \enspace .
\end{equation}
In the recent extension to Multi-Channel Correlation Filters (MCCFs)~\cite{Boddeti2013,Henriques2013,KianiGaloogahi2013}, the matrix becomes block two-level circulant and its block-diagonalisation is directly obtained according to
\begin{equation}
\SHat_{uv} = \sum_{\x} \xhat[u, v] \, \xhat^{*}[u, v] \enspace .
\end{equation}

Henriques et al.~\cite{Henriques2013} suggested the use of MCCFs as an alternative to HNM, approximating the set of all translated windows in an image with the set of all circular shifts of a subset of windows which cover the image (refer to Figure~\ref{fig:mccf-hnm}).
In the following section, we develop an alternative method to obtain a circulant covariance matrix from its Toeplitz counterpart, eliminating the need to choose a subset of windows in each image.
Crucially, this is achieved by formulating $\h$ from equation~\eqref{eq:circ} explicitly in the spatial domain, rather than formulating $\SHat$ directly in the Fourier domain.

Solving LDA with this circulant matrix is equivalent to obtaining a detector using MCCF, since correlation filters are merely an efficient method to solve least-squares regression, and a famous result states that least-squares regression is identical to LDA when the desired outputs take on exactly two distinct values~\cite{Fukunaga1990}.
While correlation filters do permit a desired response to be specified for each individual circular shift, it was empirically found in~\cite{Henriques2013} that an impulse response (1 at the correct location and 0 everywhere else) gave the best performance.

Whereas the approach of Henriques et al.~\cite{Henriques2013} still necessitates re-traversal of the negative set to train a detector of a different size, our approach inherits the attribute of only needing to traverse it once ever.

\subsection{From Toeplitz to circulant}

This section formulates an expression for the elements of the circulant matrix $h_{pq}[du, dv]$ from those of the Toeplitz matrix $g_{pq}[du, dv]$.
This is performed in the same way that MCCF obtains a circulant matrix: by incorporating all circular shifts of all signals in a set.
The set which we consider is one which produces a Toeplitz matrix.

\begin{figure}[t]
\centering
\includegraphics[width=90mm]{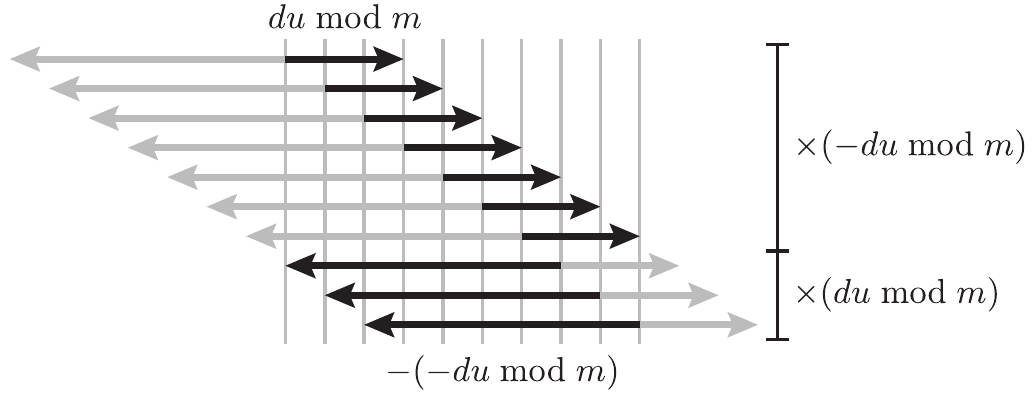}
\caption{
Under periodic extension, a relative displacement $du \ne 0$ from every position in the signal is more often observed as the smaller displacement of the two modulo complements.
For example, a small positive displacement and a large negative displacement are both predominantly observed as a small positive displacement.
}
\label{fig:mod}
\end{figure}

Let us first consider one-dimensional (but still multi-channel) signals of length $m$.
Recall that \cite{Hariharan2012} assumes that the set of examples $\mathcal{X}$ from which the statistics are estimated results in a covariance matrix with Toeplitz structure
\begin{equation}
S_{(u, p), (i, q)} \; = \; \sum_{\x \in \mathcal{X}} x_{p}[u] \, x_{q}[i] \; = \; g_{pq}[i-u] \enspace .
\label{eq:covar-toep}
\end{equation}
To obtain a circulant covariance matrix, instead consider the statistics of the augmented set containing all circular shifts $\tau = 0, \dots, m-1$ of every example
\begin{equation}
S_{(u, p), (i, q)} = \frac{1}{m} \sum_{\x \in \mathcal{X}} \sum_{\tau = 0}^{m-1} x_{p}[(\tau+u) \bmod m] \, x_{q}[(\tau+i) \bmod m] \enspace .
\end{equation}
This is shown to be circulant by making the substitution $\tau \leftarrow \tau - u$
\begin{equation}
S_{(u, p), (i, q)}
\; = \; \frac{1}{m} \sum_{\x \in \mathcal{X}} \sum_{\tau = 0}^{m-1} x_{p}[\tau] \, x_{q}[(\tau+i-u) \bmod m]
\; = \; h_{pq}[(i-u) \bmod m]
\label{eq:covar-circ-shifts}
\end{equation}
since $[a + (b \bmod m)] \bmod m = (a + b) \bmod m$.
To obtain $h$ from $g$, we introduce $du = i-u$ and split the summation based on whether $(\tau + du) \bmod m \ge \tau$.
Thus the inner sum in the above expression becomes
\begin{align}
\sum_{\tau = 0}^{m-1} x_{p}[\tau] \, x_{q}[(\tau+du) \bmod m]
= \sum_{\tau = 0}^{(-du \bmod m) - 1} x_{p}[\tau] \, x_{q}[\tau+(du \bmod m)] \nonumber \\
+ \sum_{\tau = (-du \bmod m)}^{m-1} x_{p}[\tau] \, x_{q}[\tau-(-du \bmod m)] \enspace .
\label{eq:split-sum-1d}
\end{align}
Combining equations~\eqref{eq:covar-toep}, \eqref{eq:covar-circ-shifts} and~\eqref{eq:split-sum-1d}, we obtain the final formula for the one-dimensional case
\begin{equation}
h_{pq}[du] \; = \; (1-\theta) \, g_{pq}[du \bmod m] + \theta \, g_{pq}[-(-du \bmod m)]
\label{eq:convex-1d}
\end{equation}
with $\theta = (du \bmod m) / m$.
This is a convex combination of the Toeplitz covariance for the relative displacements of $(du \bmod m)$ and $-(-du \bmod m)$, with greater weight given to the smaller of the two.
The intuition behind this is that, under periodic extension, a given displacement from every position in the signal is more often observed as the shorter displacement of its two modulo complements (see Figure~\ref{fig:mod}).

The case for 2D signals is more involved since displacements can wrap around horizontal and/or vertical boundaries.
Elements of the circulant matrix are given
\begin{alignat}{5}
h_{pq}[du, dv]
& = \;       & (1 - \alpha) && (1 - \beta) & \, g_{pq}[ & du \bmod m,     && \; dv \bmod n] \nonumber \\
& \quad + \; & (1 - \alpha) && \beta       & \, g_{pq}[ & du \bmod m,     && \; -(-dv \bmod n)] \nonumber \\
& \quad + \; & \alpha       && (1 - \beta) & \, g_{pq}[ & -(-du \bmod m), && \; dv \bmod n] \nonumber \\
& \quad + \; & \alpha       && \beta       & \, g_{pq}[ & -(-du \bmod m), && \; -(-dv \bmod n)]
\label{eq:convex-2d}
\end{alignat}
with $\alpha = (du \bmod m) / m$, $\beta = (dv \bmod n) / n$.

\subsection{Correlation filters as preconditioners}

\begin{figure}
\centering
\includegraphics[scale=0.8]{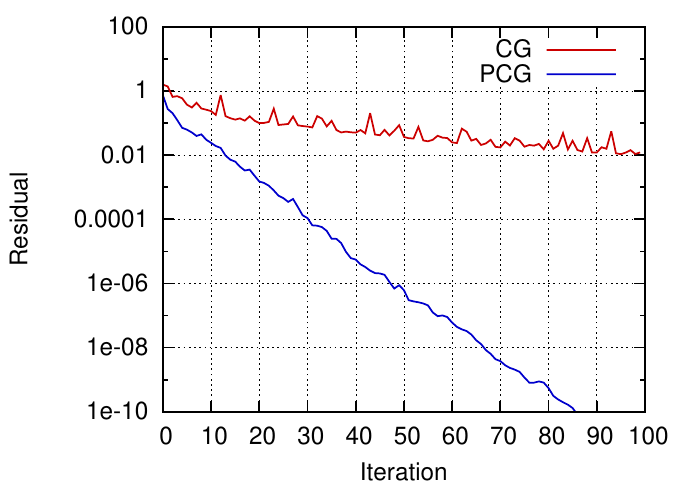}
\caption{
Convergence of conjugate gradient versus preconditioned conjugate gradient using the circulant matrix.
Residual is measured as $\|\SMat \w - \bvec\| / \|\bvec\|$.
}
\label{fig:pcg-converge}
\end{figure}

The circulant matrices described in equations~\eqref{eq:convex-1d} and~\eqref{eq:convex-2d} are each in fact the nearest, in the Frobenius sense, to the Toeplitz matrix from which they were derived.
These are the multi-channel two-level analogue of the matrices which Chan~\cite{TChan1988} proposed to use as a preconditioner.
Despite the negative theoretical result of~\cite{SerraCapizzano2000}, we find that this preconditioner results in significantly faster convergence in practice, as shown in Figure~\ref{fig:pcg-converge}.

To summarise, this leaves us with several options to learn a detector.
Firstly, we can choose to solve either the Toeplitz or the circulant system.
If we choose to solve the Toeplitz system, then we can either solve it directly by Cholesky decomposition or use CG, with or without the circulant matrix as a preconditioner.
The practical time and memory demands of these algorithms are presented in Section~\ref{sec:results}.

\subsection{Related work}

A number of previous works have employed the FFT during training.
Anguita et al.~\cite{Anguita2000} used it to efficiently compute subgradients when training an SVM across all windows in a set of images.
However, this is even slower than HNM (which uses the FFT when searching for false positives), since the negative set must be traversed per gradient descent iteration.
Stochastic subgradient descent has been used to accelerate the training of an SVM within HNM~\cite{Wijnhoven2010}, although this does nothing to alleviate the burden of searching the negative set.
Dubout and Fleuret~\cite{Dubout2013} treated images as mini-batches within stochastic descent and used the FFT to efficiently compute the subgradient of the objective function across all windows in an image.
However, this still requires access to the explicit negative examples.
Rodriguez et al.~\cite{Rodriguez2013} propose an objective function which contains both hinge loss and correlation filter terms.
This provides a boost in classification performance but is no faster to train than an SVM.
Henriques et al.~\cite{Henriques2013} considered Support Vector Regression (SVR) in addition to the least-squares problem.
However, their formulation depended on the contentious assumption that the $L_{1}$ norm of a vector is well approximated by the $L_{1}$ norm of its Fourier transform.
We feel that this needs a more thorough examination, which is beyond the scope of this paper.

\section{Empirical results}
\label{sec:results}

The proposed approaches were evaluated on the task of pedestrian detection in the INRIA dataset~\cite{Dalal2005}.
This dataset was chosen because it is well-understood and evaluation is relatively straightforward, to avoid the conflation of different aspects of the task.
Remember that our goal is to show not state-of-the-art performance but comparable performance obtained in a fraction of the time.
The statistics were estimated from a significant subset of ImageNet~\cite{Deng2009} to demonstrate that our algorithm works at scale.

\subsection{Detector quality}

\begin{figure}[t]
\centering
\includegraphics[scale=0.8]{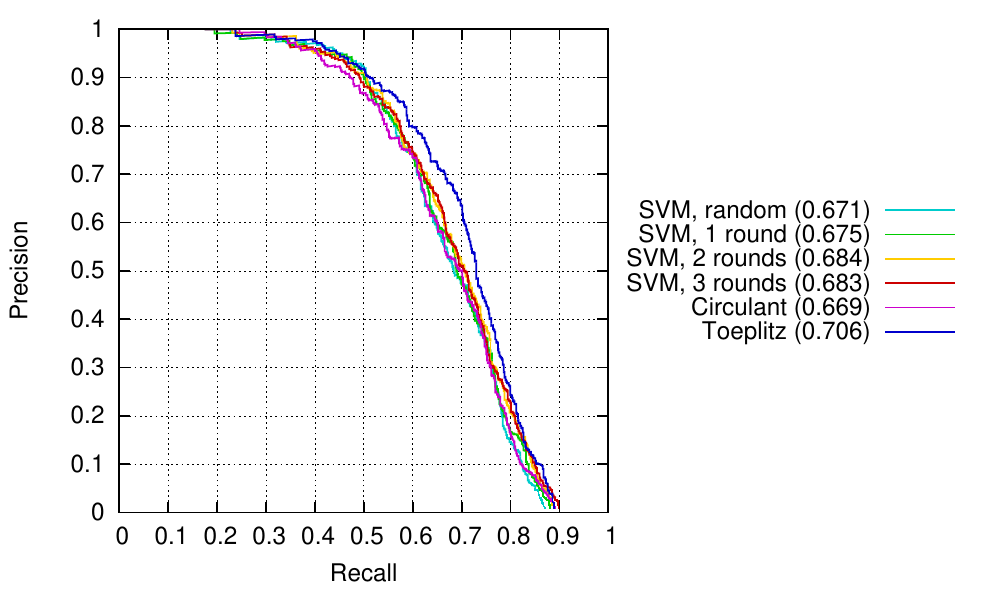}
\caption{
Precision-recall curve (incorporating multi-scale search and non-maxima suppression) for HNM and our two distinct methods.
}
\label{fig:roc}
\end{figure}

Figure~\ref{fig:roc} compares the performance of our methods to several rounds of hard negativing mining.
We find that a detector trained using circulant statistics is comparable to that learnt by an SVM on a random set of examples, and a detector trained using Toeplitz statistics is comparable to that learnt using HNM.
It's critical to remember, however, that the structured covariance methods do not need to load any negative examples into memory.

\subsection{Time and memory}

\begin{figure}[t]
\centering
\includegraphics[scale=0.75]{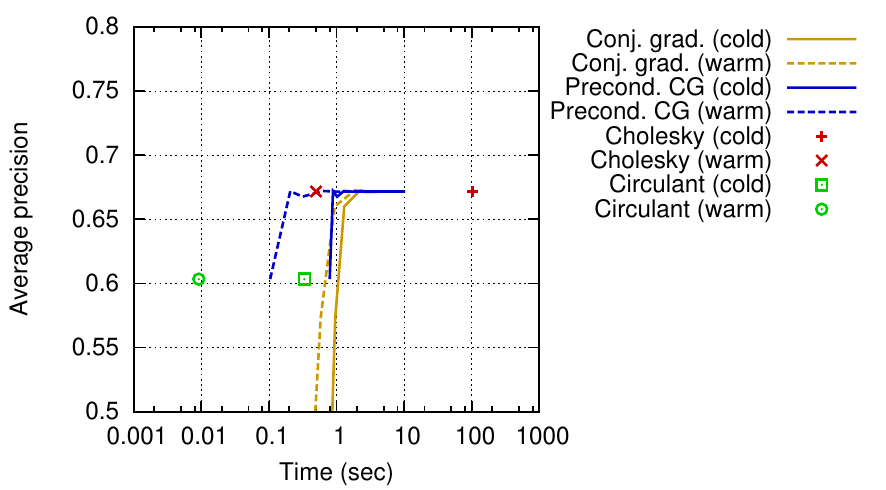}
\hspace{1em}
\includegraphics[scale=0.75]{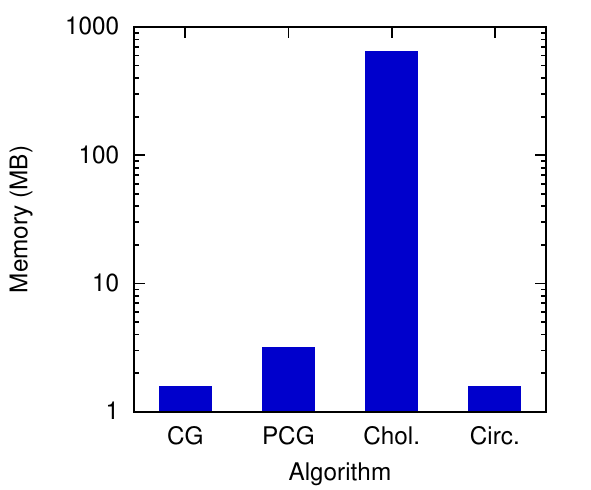}
\caption{
Left:
Average precision (on the test set) versus training time.
Iterative methods are represented by a path and closed-form methods are points.
Compared to computing a Cholesky factorisation, the conjugate gradient methods offer nearly two orders of magnitude improvement.
The pure circulant approach can be used to obtain a slightly worse detector in fractions of a second.
``Cold'' times include factorisation and pre-computable FFTs and ``warm'' times do not.
All methods must start cold per distinct template size.
Right:
Theoretical memory requirement of each learning algorithm (note the log scale).
The large size of Cholesky factorisations makes it impractical to cache factorisations for more than a few sizes.
}
\label{fig:time-mem}
\end{figure}

The key claims of this paper are evidenced in Figure~\ref{fig:time-mem}.
We present ``cold'' and ``warm'' times which include and exclude respectively pre-computable factorisations and transforms.
All methods must start cold per distinct template size.
While Cholesky factorisation offers a reasonably fast solution given cached factors, this data is prohibitively large (hundreds of MB) for moderate template sizes.
Therefore, for applications where detectors of heterogeneous sizes must be computed, huge computational gains are available.
It is much more elegant to be able to solve these systems in a modest memory footprint.

\subsection{Configuration}

The HOG descriptor~\cite{Dalal2005} was employed in all experiments, borrowing the implementation of~\cite{Felzenszwalb2010Object}, although we removed a few cells to eliminate boundary artefacts.
The spatial binning and downsampling parameter was chosen at 4 pixels.
This yielded a $12 \times 28$ feature template with 31 channels (a total of 13,020 dimensions) from a $68 \times 132$-pixel image centred about a $32 \times 96$-pixel pedestrian.
All algorithms are our own implementation, making use of \texttt{liblinear}, \texttt{FFTW} and \texttt{LAPACK}.
Images were searched at roughly 10 scales per octave (geometric steps of 1.07).
Detections were selected greedily by score, with each detection suppressing all candidates which it either covered by more than 60\% or with which it shared an intersection-over-union of more than 30\%.
Candidates which were not a maximum in their local four-connected neighbourhood were not considered.
Following~\cite{Dalal2005} and~\cite{Henriques2013}, we only mine for difficult examples in the negative images.
An initial random set of 24k images was used ($10\times$ the size of the positive set), with 2.4k more hard negatives added in each round.
The covariance and mean were gathered from four million random images in ImageNet~\cite{Deng2009}.
However, experiments suggest that there is no discernible difference to using 64k images, and initial experiments suggest 1000 or even 100 images is enough to achieve similar performance.
In all experiments we added $\lambda \mathbf{I}$ to the covariance matrix with $\lambda = 10^{-4}$.
Detections were deemed to be true positives if they have more than 50\% intersection-over-union overlap with a ground-truth box.
Each ground truth label can only match to one detection and this is performed greedily, with the highest scoring detection taking the rectangle with which it overlaps the most.

\section{Conclusion}

This paper has investigated the use of structured covariance matrices to rapidly learn object detectors from a large negative set.
Compared to existing methods employing Toeplitz structure, identical detectors are obtained in orders of magnitude less time and memory.
It has also presented a method to derive a Multi-Channel Correlation Filter from the stationary covariance matrix, alleviating the need to sample explicit examples during training.
For a slight degradation of performance, this avenue offers a further order of magnitude increase in speed.
These results are exciting for any applications which employ linear templates in a sliding window context, but in particular for those which either need to learn templates on the fly or to learn a multitude of templates.

\bibliographystyle{splncs03}
\bibliography{library}

\end{document}